\documentclass{article} 
\usepackage{iclr2015,times}
\usepackage{hyperref}
\usepackage{amsmath}
\usepackage{amssymb}
\usepackage{amsfonts}
\usepackage{url}
\usepackage{graphicx}
\newcommand{\bm}[1]{\mathbf #1}
\newcommand{\aaa}{\bm{a}}
\newcommand{\relu}{\mathrm{ReLU}}

\newcommand{\bdelta}{\boldsymbol{\delta}}
\newcommand{\hh}{\bm{h}}

\newcommand{\WW}{\bm{W}}
\newcommand{\bb}{\bm{b}}

\newcommand{\EE}{\bm{E}}
\newcommand{\del}{\partial}
\newcommand{\dEda}{\frac{\boldsymbol{\del} \EE}{\boldsymbol{\del} \aaa}}

\title{Random Walk Initialization for Training Very Deep Feedforward Networks}

\author{
David Sussillo \\
Google Inc. \\
Mountain View, CA, 94303, USA \\
\texttt{sussillo@google.com} \\
\And
L.F. Abbott \\
Departments of Neuroscience and Physiology and Cellular Biophysics\\
Columbia University\\
New York, NY, 10032, USA \\
\texttt{lfabbott@columbia.edu} \\
}

\iclrconference 

\begin{document}

\maketitle

\begin{abstract}
Training very deep networks is an important open problem in machine learning.  One of many difficulties is that the norm of the back-propagated error gradient can grow or decay exponentially.  Here we show that training very deep {\it feed-forward} networks (FFNs) is not as difficult as previously thought.  Unlike when back-propagation is applied to a recurrent network, application to an FFN amounts to multiplying the error gradient by a different random matrix at each layer.  We show that the successive application of correctly scaled random matrices to an initial vector results in a random walk of the log of the norm of the resulting vectors, and we compute the scaling that makes this walk unbiased.  The variance of the random walk grows only linearly with network depth and is inversely proportional to the size of each layer.  Practically, this implies a gradient whose log-norm scales with the {\it square root} of the network depth and shows that the vanishing gradient problem can be mitigated by increasing the width of the layers.  Mathematical analyses and experimental results using stochastic gradient descent to optimize tasks related to the MNIST and TIMIT datasets are provided to support these claims.  Equations for the optimal matrix scaling are provided for the linear and $\relu$ cases.  
\end{abstract}

\section{Introduction}

Since the early 90s, it has been appreciated that deep neural networks suffer from a vanishing gradient problem \citep{Hochreiter91-small}, \citep{Bengio_icnn93}, \citep{Bengio-trnn93}, \citep{hochreiter2001gradient}.  The term vanishing gradient refers to the fact that in a feedforward network (FFN) the back-propagated error signal typically decreases (or increases) exponentially as a function of the distance from the final layer.  This problem is also observed in recurrent networks (RNNs), where the errors are back-propagated in time and the error signal decreases (or increases) exponentially as a function of the  distance back in time from the current error.  Because of the vanishing gradient, adding many extra layers in FFNs or time points in RNNs does not usually improve performance.

Although it can be applied to both feedforward and recurrent networks, the analysis of the vanishing gradient problem is based on a recurrent architecture (e.g. \citep{Hochreiter91-small}).  In a recurrent network, back-propagation through time involves applying similar matrices repeatedly to compute the error gradient.  The outcome of this process depends on whether the magnitudes of the leading eigenvalues of these matrices tend to be greater than or less than one\footnote{Excluding highly non-normal matrices.}.  Eigenvalue magnitudes greater than one produce exponential growth, and less than one produces exponential decay.  Only if the magnitude of the leading eigenvalues are tightly constrained can there be a useful ``non-vanishing" gradient.  Although this fine-tuning can be achieved by appropriate initialization, it will almost surely be lost as the optimization process goes forward.

Interestingly, the analysis is very different  for an FFN with randomly initialized matrices at each layer.  When the error gradient is computed in a FFN, a different matrix is applied at every level of back-propagation.  This small difference can result in a wildly different behavior for the magnitude of the gradient norm for FFNs compared to RNNs.  Here we show that correctly initialized FFNs suffer from the vanishing gradient problem in a far less drastic way than previously thought, namely that the magnitude of the gradient  scales only as the square root of the depth of the network. 

Different approaches to training deep networks (both feedforward and recurrent) have been studied and applied, such as pre-training \citep{hinton2006reducing}, better random initial scaling \citep{glorot2010understanding},\citep{sutskever2013importance}, better optimization methods \citep{martens2010deep}, specific architectures \citep{krizhevsky2012imagenet}, orthogonal initialization \citep{saxe2013exact}, etc.  Further, the topic of why deep networks are difficult to train is also an area of active research \citep{glorot2010understanding}, \citep{pascanu2012difficulty}, \citep{saxe2013exact}, \citep{pascanu2014saddle}.  

Here, we address the vanishing gradient problem using mathematical analysis and computational experiments that study the training error optimized deep-networks.  We analyze the norm of vectors that result from successive applications of random matrices, and we show that the analytical results hold empirically for the back-propagation equations of nonlinear FFNs with hundreds of layers.  We present and test a basic heuristic for initializing these networks, a procedure we call \textbf{Random Walk Initialization} because of the random walk of the log of the norms (log-norms) of the back-propagated errors.

\section {Analysis and Proposed Initialization}

\subsection{The Magnitude of the Error Gradient in FFNs}
We focus on feedforward networks of the form
\begin{align}
  \aaa_{d} &= g\WW_{d}\;\hh_{d-1} + \bb_{d} \label{eq:fp1}\\
  \hh_{d} &= f\left(\; \aaa_{d}\;\right), \label{eq:fp2}
\end{align}
with $\hh_d$ the vector of hidden activations, $\WW_d$ the linear transformation, and $\bb_d$ the biases, all at depth $d$, with $d = 0, 1, 2, \ldots, D$.  The function $f$ is an element-wise nonlinearity that we will normalize through the derivative condition $f'(0) = 1$, and $g$ is a scale factor on the matrices.  We assume that the network has $D$ layers and that each layer has width $N$ (i.e. $\hh_d$ is a length $N$ vector).  Further we assume the elements of $\WW_d$ are initially drawn i.i.d.\ from a Gaussian distribution with  zero mean and variance $1/N$.  Otherwise the elements are set to 0.  The elements of $\bb_d$ are initialized to zero.  We define $\hh_0$ to be the inputs and $\hh_D$ to be the outputs. 

We assume that a task is defined for the network by a standard objective function, $E$.  Defining $\bdelta_d \equiv \left.\dEda\right|_d$,  the corresponding back-propagation equation is
\begin{align}
  \bdelta_d &= g \tilde\WW_{d+1} \; \bdelta_{d+1}, \label{eq:bp} 
\end{align}
where $\tilde\WW_d$ is a matrix with elements given by 
\begin{align}
  \tilde\WW_d(i, j) &= f'\!\left(a_{d}(i)\right) W_{d}(j, i) \label{eq:tW}.
\end{align}
The evolution of the squared magnitude of the gradient vector,  $|\bdelta_d|^2$, during back-propagation can be written as
\begin{align}
|\bdelta_d|^2 &= g^2z_{d+1}| \bdelta_{d+1}|^2,
\end{align}
where we have defined, for reasons that will become apparent,
\begin{align}
z_d &= \left|\tilde\WW_d\bdelta_d/|\bdelta_d|\right|^2 .  \label{eq:zDef} 
\end{align}
The entire evolution of the gradient magnitude across all $D$ layers of the network is then described by
\begin{align}
Z = \frac{|\bdelta_0|^2}{|\bdelta_{D}|^2} &= g^{2D}\prod_{d=1}^Dz_{d} ,  \label{eq:ZDef} 
\end{align}
where we have defined the across-all-layer gradient magnitude ratio as $Z$.  Solving the vanishing gradient problem amounts to keeping $Z$ of order 1, and our proposal is to do this by appropriately adjusting $g$.  Of course, the matrices $\WW$ and $\tilde\WW$ change during learning, so we can only do this for the initial configuration of the network before learning has made these changes.  We will discuss how to make this initial adjustment and then show experimentally that it is sufficient to maintain useful gradients even during learning.

Because the matrices $\WW$ are initially random, we can think of the $z$ variables defined in equation (\ref{eq:zDef}) as random variables.  Then, $Z$, given by equation (\ref{eq:ZDef}), is proportional to a product of random variables and so, according to the central limit theorem for products of random variables, $Z$ will be approximately log-normal distributed for large $D$.  This implies that the distribution for $Z$ is long-tailed.  For applications to neural network optimization, we want a procedure that will regularize $Z$ in most cases, resulting in most optimizations making progress, but are willing to tolerate the occasional pathological case, resulting in a failed optimization.  This means that we are not interested in catering to the tails of the $Z$ distribution.  To avoid issues associated with these tails, we choose instead to consider the logarithm of equation (\ref{eq:ZDef}),
\begin{align}
\ln(Z) &= D\ln(g^2) + \sum_{d=1}^D\ln(z_{d}) .  \label{eq:LogZDef} 
\end{align}
The sum in this equation means that we can think of $\ln(Z)$ as being the result of a random walk, with  step $d$ in the walk given by the random variable $\ln(z_d)$.  The goal of Random Walk Initialization is to chose $g$ to make this walk unbiased.  Equivalently, we choose $g$ to make $\ln(Z)$ as close to zero as possible.  

\subsection{Calculation of the Optimal $g$ Values}

Equation (\ref{eq:LogZDef}) describes the evolution of the logarithm of the error-vector norm as the output-layer vector $\bdelta_D$ is back-propagated through a network.  In an actual application of the back-propagation algorithm,  $\bdelta_D$ would be computed by propagating an input forward through the network and comparing the network output to the desired output for the particular task being trained.  This is what we will do as well in our neural network optimization experiments, but we would like to begin by studying the vanishing gradient problem in a broader context, in particular, one that allows a general discussion independent of the particular task being trained.  To do this, we study what happens to randomly chosen vectors $\bdelta_D$ when they are back-propagated, rather than studying specific vectors that result from forward propagation and an error computation.  Among other things, this implies that the $\bdelta_D$ we use are uncorrelated with the $\WW$ matrices of the network.  Similarly, we want to make analytic statements that apply to all networks, not one specific network.  To accomplish this, we average over realizations of the matrices $\tilde\WW$ applied during back-propagation.  After presenting the analytic results, we will show that they provide excellent approximations when applied to back-propagation calculations on specific networks being trained to perform specific tasks. 

When we average over the $\tilde\WW$ matrices, each layer of the network becomes equivalent, so we can write
\begin{align}
\langle\ln(Z)\rangle &= D\left(\ln(g^2) + \langle\ln(z)\rangle\right) = 0, 
\end{align}
determining the critical value of $g$ as
\begin{align}
g &= \exp\left(-\frac{1}{2}\langle\ln(z)\rangle\right)  . \label{eq:gOpt}
\end{align}
Here $z$ is a random variable determined by
\begin{align}
z &= \left|\tilde\WW\bdelta/|\bdelta|\right|^2 ,  \label{eq:zEff}
\end{align}
with $\tilde\WW$ and $\bdelta$ chosen from the same distribution as the $\tilde\WW_d$ and $\bdelta_d$ variables of the different layers of the network (i.e. we have dropped the $d$ index).

We will compute the optimal $g$ of equation (\ref{eq:gOpt}) under the assumption that $\tilde\WW$ is i.i.d.\ Gaussian with zero mean.  For linear networks, $f' = 1$ so $\tilde\WW = \WW^T$, and this condition is satisfied due to the definition of $\WW$.  For the $\relu$ nonlinearity, $f'$ effectively zeros out approximately half of the rows of $\tilde\WW$, but the Gaussian assumption applies to the non-zero rows.  For the $\tanh$ nonlinearity, we will rely on numerical rather than analytic results and thus will not need this assumption.

When $\tilde\WW$ is Gaussian, so is $\tilde\WW\bdelta/|\bdelta|$, independent of the distribution or value of the vector $\bdelta$ (that is, the product of a Gaussian matrix and a unit vector is a Gaussian vector).  The fact that $\tilde\WW\bdelta/|\bdelta|$ is Gaussian for any vector $\bdelta$ means that we do not need to consider the properties of  the $\bdelta_d$ vectors at different network layers, making the calculations much easier.  It also implies that $z$ is $\chi^2$ distributed because it is the squared magnitude of a Gaussian vector.  More precisely, if the elements of the $N\times N$ matrix $\tilde\WW$ have variance $1/N$, we can write $z = \eta/N$, where $\eta$ is distributed according to $\chi^2_N$, that is, a $\chi^2$ distribution with $N$ degrees of freedom. 

Writing $z = \eta/N$, expanding the logarithm in a Taylor series about $z =1$ and using the mean and variance of the distribution $\chi^2_N$, we find
\begin{align}
\left\langle\ln(z)\right\rangle \approx \left\langle (z-1)\right\rangle - \frac{1}{2}\left\langle (z-1)^2\right\rangle = - \frac{1}{N} .
\end{align}
From equation (\ref{eq:gOpt}), this implies, to the same degree of approximation, that the optimal $g$ is
\begin{align}
  g_{\mbox{\footnotesize{linear}}} = \exp\left(\frac{1}{2N}\right).  \label{eq:gOptComput}
\end{align}
The slope of the variance of the random walk of $\ln(Z)$ is given to this level of approximation by  
\begin{align}
\Big\langle \left(\ln(z)\right)^2 \Big\rangle - \Big\langle\ln(z)\Big\rangle^2 = \frac{1}{2N}.  \label{eq:rwv}
\end{align}
Note that this is inversely proportional to $N$.  These expressions are only computed to lowest order in a $1/N$ expansion, but numerical studies indicate that they are reasonably accurate (more accurate than expressions that include order $1/N^2$ terms) over the entire range of $N$ values.  

For the $\relu$ case, $\tilde\WW \neq \WW^T$, but it is closely related because the factor of $f'$, which is equal to either 0 or 1, sets a fraction of the rows of $\tilde\WW$ to 0 but leaves the other rows unchanged.  Given the zero-mean initialization of $\WW$, both values of $f'$ occur with probability 1/2.  Thus the derivative of the $\relu$ function sets $1-M$ rows of $\tilde\WW$ to 0 and leaves $M$ rows with Gaussian entries.  The value of $M$ is drawn from an $N$ element binomial distribution with $p = 1/2$, and $z$ is the sum of the squares of $M$ random variables with variance $1/N$.  We write $z = \eta/N$ as before, but in this case, $\eta$ is distributed according to $\chi^2_M$.  This means that $z$ is a doubly stochastic variable: first $M$ is drawn from the binomial distribution and then $\eta$ is drawn from $\chi^2_M$.  Similarly, the average $\langle\ln(z)\rangle$ must now be done over both the $\chi^2$ and binomial distributions.  A complication in this procedure, and in using $ \relu$ networks in general, is that if $N$ is too small (about $N < 20$) a layer may have no activity in any of its units.  We remove these cases from our numerical studies and only average over nonzero values of $M$.

We can compute $\langle\ln(z)\rangle$ to leading order in $1/N$ using a Taylor series as above, but expanding around $z = 1/2$ in this case, to obtain
\begin{align}
\langle\ln(z)\rangle \approx -\ln(2) - \frac{2}{N} .
\end{align}
However, unlike in the linear case, this expression is not a good approximation over the entire $N$ range.  Instead, we computed $\langle\ln(z)\rangle$ and $\langle(\ln(z))^2\rangle$ numerically and fit simple analytic expressions to the results to obtain
\begin{align}
\langle\ln(z)\rangle \approx -\ln(2) - \frac{2.4}{\max(N, 6) - 2.4} , \label{eq:reLULog}
\end{align}
and
\begin{align}
\Big\langle \left(\ln(z)\right)^2 \Big\rangle - \Big\langle\ln(z)\Big\rangle^2  \approx \frac{5}{\max(N, 6) - 4} .
\end{align}
From equation (\ref{eq:reLULog}), we find
\begin{align}
g_{\mbox{\footnotesize{ReLU}}} = \sqrt{2}\exp\left(\frac{1.2}{\max(N, 6)-2.4}\right) .  \label{eq:gReLU}
\end{align}

Computing these averages with the $\tanh$ nonlinearity is more difficult and, though it should be possible, we will not attempt to do this.  Instead, we report numerical results below.  In general, we should expect the optimal $g$ value for the $\tanh$ case to be greater than $g_{\mbox{\footnotesize{linear}}}$, because the derivative of the $\tanh$ function reduces the variance of the rows of $\tilde\WW$ compared with those of $\WW$, but less than g$_{\mbox{\footnotesize{ReLU}}}$ because multiple rows of $\tilde\WW$ are not set to 0.

\subsection{Computational Verification}

The random walks that generate $Z$ values according to equation (\ref{eq:LogZDef}) are shown in the top panel of Figure 1 for a linear network (with random vectors back-propagated).  In this case, the optimal $g$ value, given by equation (\ref{eq:gOptComput}) was used, producing an unbiased random walk (middle panel of Figure 1).  The linear increase in the variance of the random walk across layers is well predicted by variance computed in equation (\ref{eq:rwv}).

\begin{figure}[h]
\begin{center}
  \includegraphics[width=0.6\linewidth]{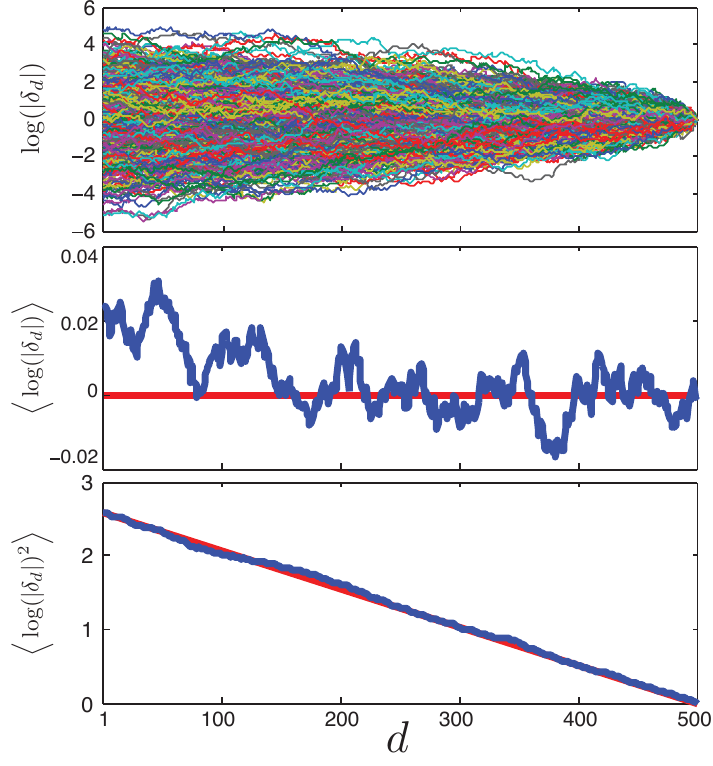}
\end{center}
\caption{Sample random walks of random vectors back-propagated through a linear network. (Top) Many samples of random walks from equation (\ref{eq:LogZDef}) with $N=100$, $D=500$ and $g=1.005$, as determined by equation (\ref{eq:gOptComput}). Both the starting vectors as well as all matrices were generated randomly at each step of the random walk.  (Middle) The mean over all instantiations (blue) is close to zero (red line) because the optimal $g$ value was used.  (Bottom) The variance of the random walks at layer $d$ (blue), and the value predicted by equation (\ref{eq:rwv}) (red). }
\end{figure}

We also explored via numerical simulation the degree to which equations (\ref{eq:gOpt}) and (\ref{eq:gReLU}) were good approximations of the dependence of $g$ on $N$.  The results are shown in Figure 2.  The top row of Figure 2 shows the predicted $g$ value as a function of the layer width, $N$, and the nonlinearity.  Each point is averaged over 200 random networks of the form given by equations (\ref{eq:fp1}-\ref{eq:bp}) with $D=200$ and both $\hh_0$ and $\bdelta_D$ set to a random vectors whose elements have unit variance.  The bottom row of Figure 2 shows the growth of the magnitude of $\bdelta_0$ in comparison to $\bdelta_D$ for a fixed $N=100$, as a function of the $g$ scaling parameter and the nonlinearity.  Each point is averaged over 400 random instantiations of equations (\ref{eq:fp1}-\ref{eq:fp2}) and back-propagated via equations (\ref{eq:bp}).  The results show the predicted optimal $g$ values from equations (\ref{eq:gOptComput}) and (\ref{eq:gReLU}) match the data well, and they provide a numerical estimate of the optimal $g$ value of the $\tanh$ case.  In addition, we see that the range of serviceable values for $g$ is larger for $\tanh$ than for the linear or $\relu$ cases due to the saturation of the nonlinearity compensating for growth due to $g\WW_d$.

\begin{figure}[h]
\begin{center}
  \includegraphics[width=\linewidth]{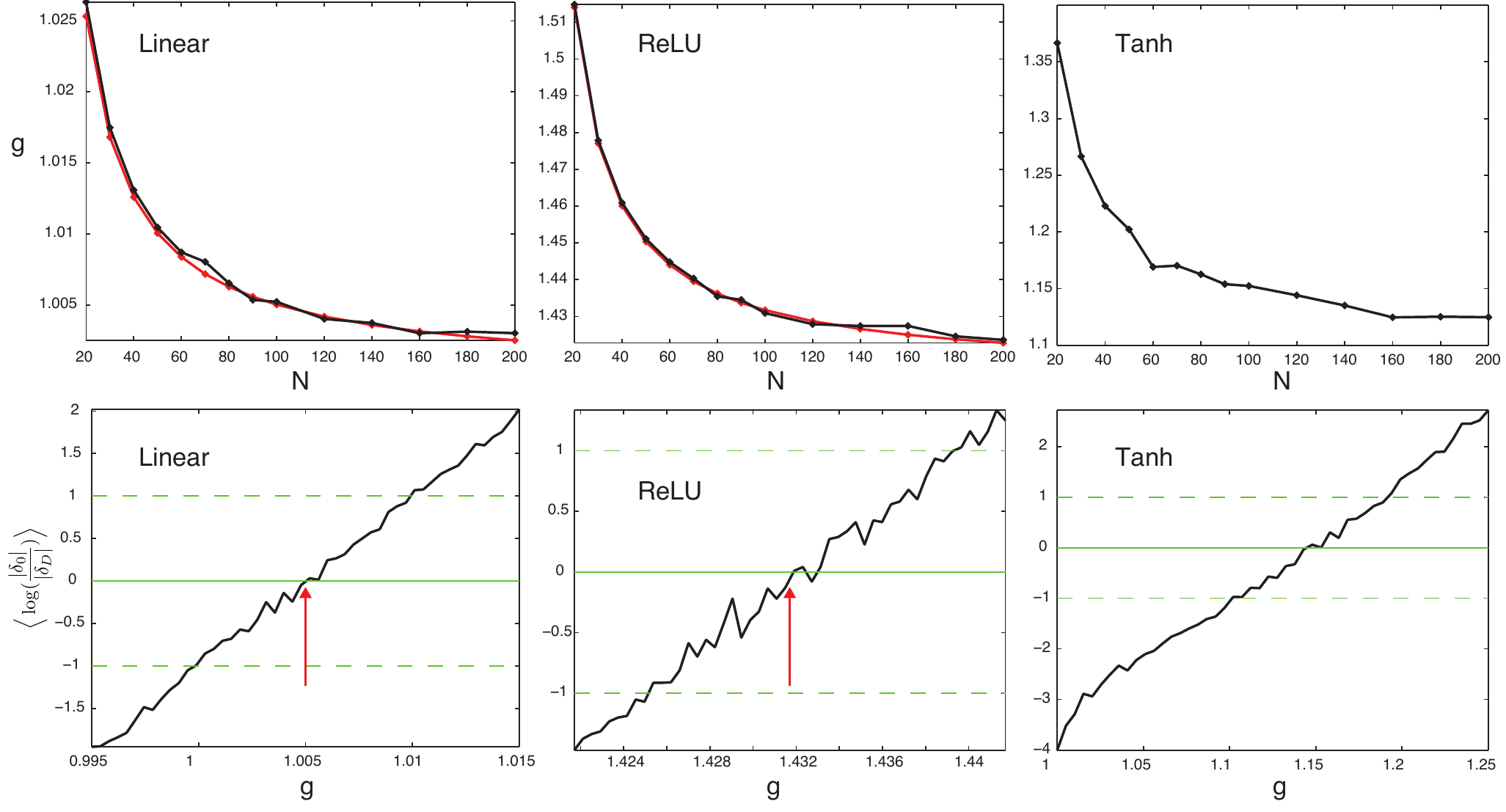}
\end{center}
\caption{Top - Numerical simulation of the best $g$ as a function of $N$, using equations (\ref{eq:fp1}-\ref{eq:bp}) using random vectors for $\hh_0$ and $\bdelta_D$. Black shows results of numerical simulations, and red shows the predicted best $g$ values from equations (\ref{eq:gOptComput}) and (\ref{eq:gReLU}). (Left) $\mbox{linear}$, (Middle) - $\relu$, (Right) - $\tanh$. Bottom - Numerical simulation of the average  $\log(|\bdelta_0|/|\bdelta_D|)$ as a function of $g$, again using equations (\ref{eq:fp1}-\ref{eq:bp}).  Results from equations (\ref{eq:gOptComput}) and (\ref{eq:gReLU}) are indicated by red arrows.  Guidelines at 0 (solid green) and -1, 1 (dashed green) are provided. }
\end{figure}

\section {Results of Training Deep Networks with Random Walk Initialization}

\subsection{Random Walk Initialization}

The general methodology used in the Random Walk Initialization is to set $g$ according to the values given in equations (\ref{eq:gOptComput}) and (\ref{eq:gReLU}) for the linear and $\relu$ cases, respectively. For $\tanh$, the values between 1.1 and 1.3 are shown to be good in practice, as shown in Figure 2 (upper right panel) and in Figure 3 (left panels).  The scaling of the input distribution itself should also be adjusted to zero mean and unit variance in each dimension. Poor input scaling will effect the back-propagation through the derivative terms in equation (\ref{eq:bp}) for some number of early layers before the randomness of the initial matrices  ``washes out'' the poor scaling.  A slight adjustment to $g$ may be helpful, based on the actual data distribution, as most real-world data is far from a normal distribution.  By similar reasoning, the initial scaling of the final output layer may need to be adjusted separately, as the back-propagating errors will be affected by the initialization of the final output layer. In summary, Random Walk Initialization requires tuning of three parameters: input scaling (or $g_1$), $g_D$, and $g$, the first two to handle transient effects of the inputs and errors, and the last to generally tune the entire network.  By far the most important of the three is $g$. 

\subsection{Experimental Methods}

To assess the quality of the training error for deep nonlinear FFNs set up with Random Walk Initialization, we ran experiments on both the MNIST and TIMIT datasets with a standard FFN defined by equations (\ref{eq:fp1}-\ref{eq:fp2}).  In particular we studied the classification problem for both MNIST and TIMIT, using cross-entropy error for multiclass classification, and we studied reconstruction of MNIST digits using auto-encoders, using mean squared error.  For the TIMIT study, the input features were 15 frames (+/- 7 frames of context, with $\Delta$ and $\Delta\Delta$).  In these studies, we focused exclusively on training error, as the effect of depth on generalization is a different problem (though obviously important) from how one can train deep FFNs in the first place.  

The general experimental procedure was to limit the number of parameters, e.g. 4e6 parameters, and distribute them between matrices and biases of each layer in a network.  The classification experiments used constant width layers, and for these experiments the actual number of parameters was the first value above the parameter limit, $p_{lim}$, such that a constant integer value of $N$ was possible.  Thus as a network got deeper, its layers also became more narrow.  For example, for the MNIST dataset, at $p_{lim}=4e6$ , for $D=4$, $N=1228$ and for $D=512$, $N=88$.  For the MNIST auto-encoder experiments, $p_{lim}=16e6$, and the code layer was 30 linear units.  The size of each layer surrounding this middle encoding layer was chosen by picking a constant increase in layer size such that the total number of parameters was first number above $p_{lim}$ that led to an integral layer width for all layers.  For example, at $D=4$, the layer sizes were $[9816\; 30\; 9816\; 784]$, while for $D=128$ the layer sizes were $[576\; 567\; 558\; ...\;  48\; 39\; 30\; 38\; 48\; ...\; 558\; 567\; 576\; 784]$.  In these auto-encoder studies we used the $\tanh$ nonlinearity, and varied the $g$ parameter per experiment, but not per layer.

Our experiments compared one depth to another so we varied the learning rates quite a bit to ensure fairness for both shallow and deep networks.  In particular, we varied the minimal and maximal learning rates per experiment.  In essence, we had an exponential learning rate schedule {\it as a function of depth}, with the minimal and maximal values of that exponential set as hyper-parameters.  More precisely, we denote the maximum depth in an experiment as $D_{max}$ (e.g. if we compared networks with depths $[4\; 8\; 16\; 32\; 64\; 128]$ in a single experiment, then $D_{max} = 128$).  Let $\lambda_{in}$ and $\lambda_{out}$ be the learning rate hyper-parameters for the input and output layers, respectively.  The exponential learning rate schedule with decay $\tau$ and scale $\alpha$, as a function of depth, took the form
\begin{align}
  \tau &= \frac{(D_{max}-1)}{\ln(\lambda_{out})-\ln(\lambda_{in})} \\
  \alpha &= \exp(\ln(\lambda_{in}) + \frac{D_{max}}{\tau})\\
  \gamma_d &= \alpha \; \exp(-\frac{D_{max}-d+1}{\tau}).
\end{align}
Then for a given network with depth $D$, potentially smaller than $D_{max}$, the learning rates were set for the actual experiment as
\begin{align}
  \lambda_{D-d} &= \gamma_{D_{max} - d}.
\end{align}
A key aspect of this learning rate scheme is that shallower networks are not overly penalized with tiny learning rates in the early layers. This is because the decay starts with layer $D$ getting learning rate $\lambda_{out}$ and goes backwards to the first layer, which gets a learning rate $\lambda_{D_{max}-D}$.  This means that for networks more shallow than $D_{max}$, $\lambda_1$ could be much larger than $\lambda_{in}$; only if $D = D_{max}$ did $\lambda_1 = \lambda_{in}$.  Some experiments had $\lambda_{in} < \lambda_{out}$, some had $\lambda_{in} > \lambda_{out}$, and we also tested the standard $\lambda_{in} = \lambda_{out}$ (no learning rate schedule as a function of depth for all experiments). For the very deep networks, varying the learning rates as a function of depth was very important, although we do not study it in depth here.  Finally, the learning rates in all layers were uniformly decayed by a multiplicative factor of 0.995 at the end of each training epoch.  

Beyond setting learning rate schedules, there were no bells and whistles.  We trained the networks using standard stochastic gradient descent (SGD) with a minibatch size of 100 for 500 epochs of the full training dataset.  We also used gradient clipping, in cases when the gradient became very large, although this was very uncommon.  The combination of hyper-parameters: the varied learning rates, depths, and $g$ values resulted in roughly 300-1000 optimizations for each panel displayed in Figure 3 and Figure 4.  

\subsection{Performance Results}

\begin{figure}[h]
\begin{center}
  \includegraphics[width=0.8\linewidth]{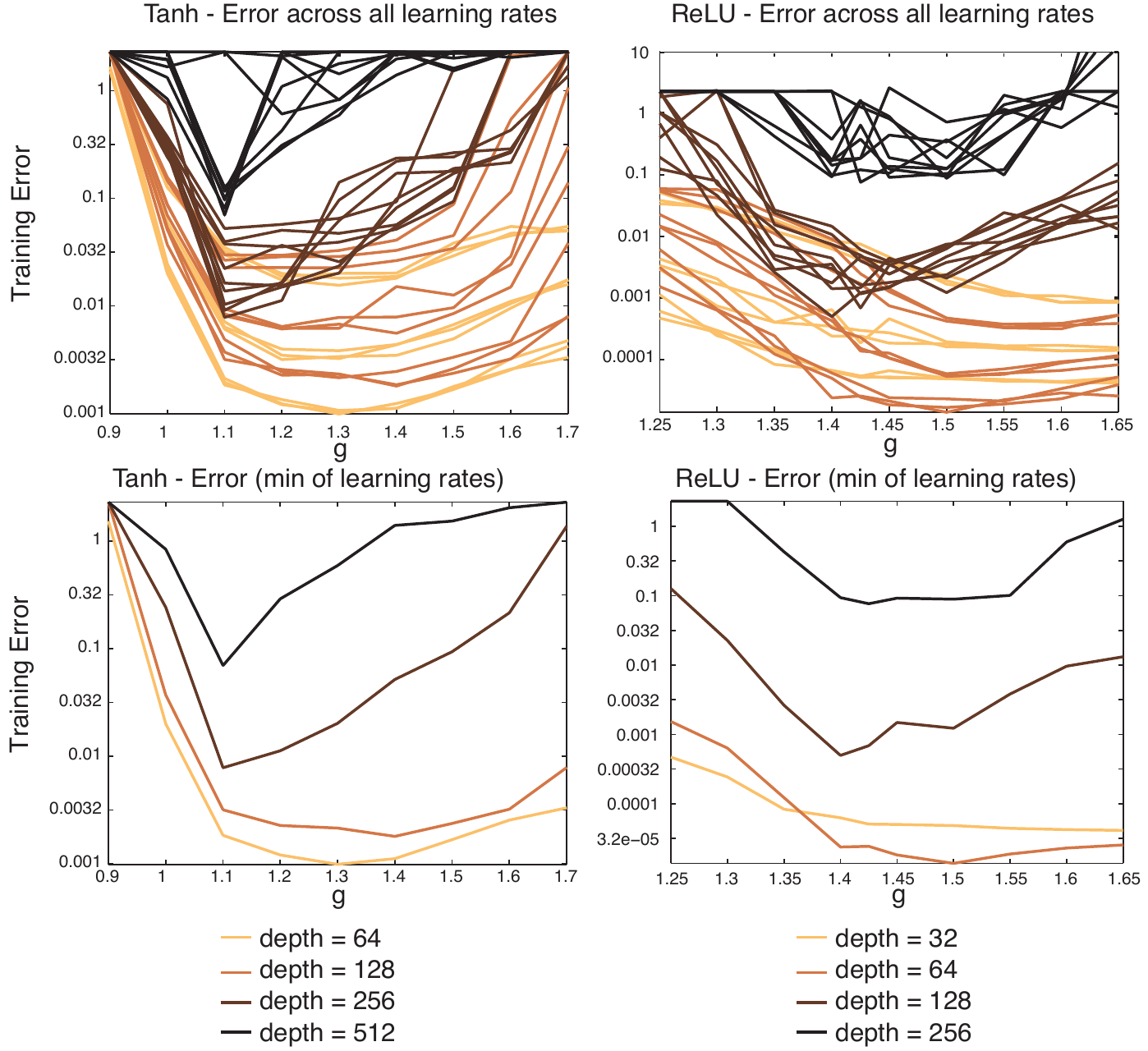}
\end{center}
\caption{Training error on MNIST as a function of $g$ and $D$.  Each simulation used a parameter limit of $4e6$.  Error shown on a $\log_{10}$ scale.  The $g$ parameter is varied on the x-axis, and color denotes various values of $D$.  (Upper left) Training error for the $\tanh$ function for all learning rate combinations.  The learning rate hyper-parameters $\lambda_{in}$ and $\lambda_{out}$ are not visually distinguished.  (Lower left) Same as upper left except showing the minimum training error for all learning rate combinations.  (Upper right and lower right) Same as left, but for the $\relu$ nonlinearity.  For both nonlinearities, the experimental results are in good agreement with analytical and experimental predictions.}
\end{figure}

We employed a first set of experiments to determine whether or not training a real-world dataset would be affected by choosing $g$ according to the Random Walk Initialization.  We trained many networks as described above on the MNIST dataset.  The results are shown in Figure 3 for both the $\tanh$ and $\relu$ nonlinearities.  Namely, for $\tanh$ the smallest training error for most depths is between $g = 1.1$ and $g=1.4$, in good agreement with Figure 2 (upper left panel).  For $\relu$ the smallest training error was between $g = 1.4$ and $g=1.55$.  These results are in very good agreement with our analytic calculations and the results shown in Figure 2 (upper middle panel).

The goal of the second set of experiments was to assess training error as a function of $D$.  In other words, does increased depth actually help to decrease the objective function?  Here we focused on the $\tanh$ nonlinearity as many believe $\relu$ is the easier function to use.  Having demonstrated the utility of correctly scaling $g$, we used a variety of $g$ values in the general optimal range above 1.  The results for MNIST classification are shown in Figure 4A.  The best training error was depth 2, with a very large learning rate.  Tied for second place, depths of 16 and 32 showed the next lowest training error.  The MNIST auto-encoder experiments are shown in Figure 4B.  Again the most shallow network achieved the best training error.  However, even depths of 128 were only roughly 2x greater in training error, thus demonstrating the effectiveness of our initialization scheme.  Mostly as a stunt, we trained networks with 1000 layers to classify MNIST.  The parameter limit was 62e6, resulting in a layer width of 249.  The results are shown in Figure 4C.  The very best networks (over all hyper-parameters) were able to achieve a performance of about 50 {\it training} mistakes.  We also tried Random Walk Initialization on the TIMIT dataset (Figure 4D) and the results were similar to MNIST.  The best training error among the depths tested was depth 16, with depth 32 essentially tied.  In summary, depth did not improve training error on any of the tasks we examined, but these experiments nevertheless provide strong evidence that our initialization is reasonable for very deep nonlinear FFNs trained on real-world data.

\begin{figure}[h]
\begin{center}
  \includegraphics[width=1.0\linewidth]{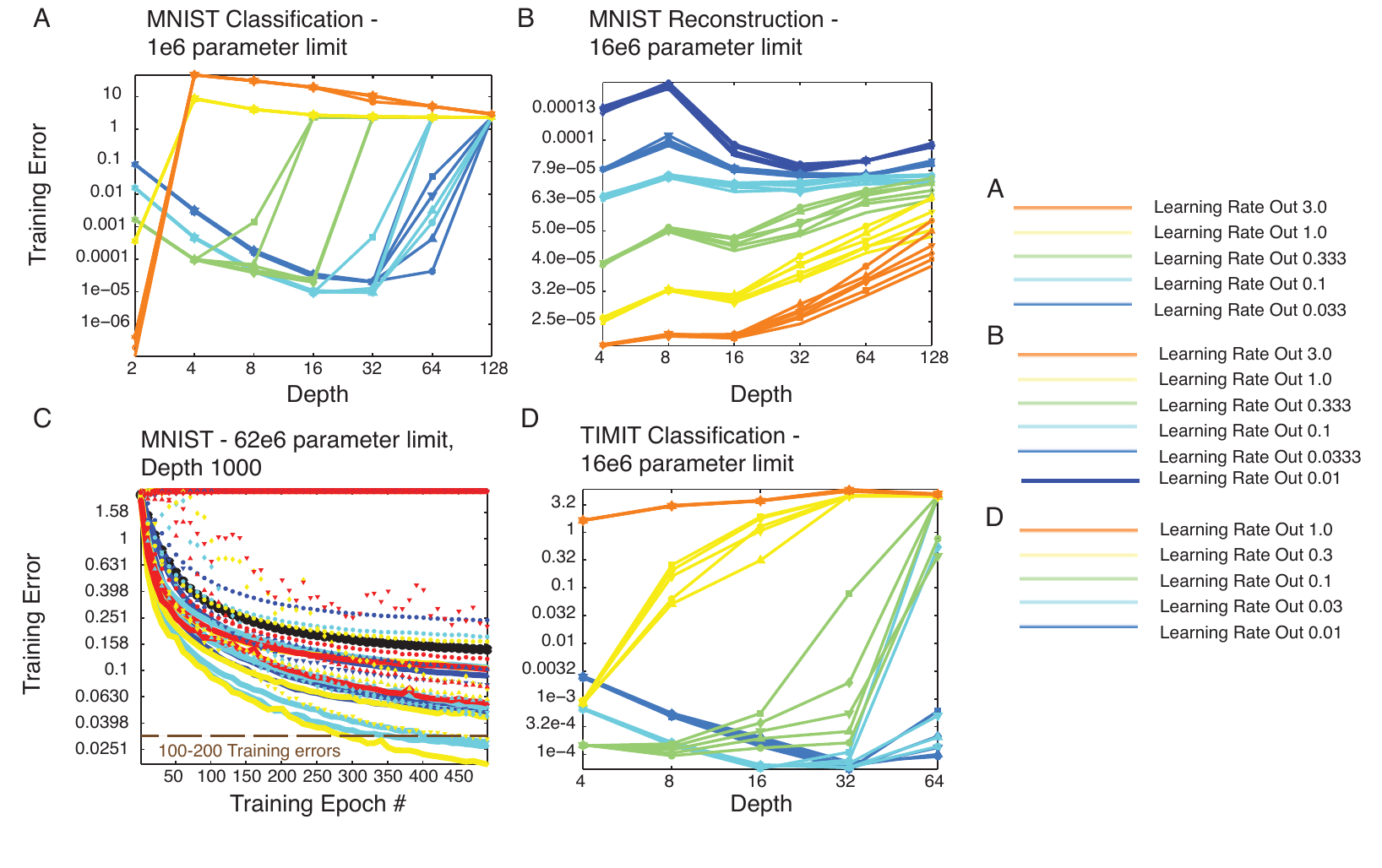}
\end{center}
\caption{Performance results using Random Walk Initialization on MNIST and TIMIT. Each network had the same parameter limit, regardless of depth.  Training error is shown on a $\log_{2}-\log_{10}$ plot. The legend for color coding of $\lambda_{out}$ is shown at right.  The values of $\lambda_{in}$ were varied with the same values as $\lambda_{out}$ and are shown with different markers.  Varied $g$ values were also used and averaged over.  For A and B, $g = [1.05, 1.1, 1.15, 1.2]$.  (A) The classification training error on MNIST as a function of $D$, $\lambda_{in}$ and $\lambda_{out}$.  (B) MNIST Auto-encoder reconstruction error as a function of hyper-parameters.  (C) Experiments on MNIST with $D = 1000$.  Training error is shown as a function of training epoch.  Hyper-parameters of $\lambda_{in}, \lambda_{out}$, were varied to get a sense of the difficulty of training such a deep network. The value of $g = 1.05$ was used to combat pathological curvature.  (D) Classification training error on TIMIT dataset.  Values of $g = [1.1, 1.15, 1.2, 1.25]$ were averaged over. }
\end{figure}

\section {Discussion}

The results presented here imply that correctly initialized FFNs, with $g$ values set as in Figure 2 or as in equation (\ref{eq:gOptComput}) for linear networks and equation (\ref{eq:gReLU}) for $\relu$ networks, can be successfully trained on real datasets for depths upwards of 200 layers.  Importantly, one may simply increase $N$ to decrease the fluctuations in the norm of the back-propagated errors.  We derived equations for the correct $g$ for both the linear and $\relu$ cases.  While our experiments explicitly used SGD and avoided regularization, there is no reason that Random Walk Initialization should be incompatible with other methods used to train very deep networks, including second-order optimization, different architectures, or regularization methods.

This study revealed a number of points about training very deep networks.  First, one should be careful with biases.  Throughout our experiments, we initialized the biases to zero, though we always allowed them to be modified.  For the most part, use of biases did not hurt the results.  However, care must be taken with the learning rates because the optimization may use the biases to quickly match the target mean across examples. If this happens, the careful initialization may be destroyed and forward progress in the optimization will cease.  Second, learning rates in very deep networks are very important. As can be seen from Figure 4D (e.g. $D = 32$), the exact learning rate scheduling made a huge difference in performance.  Third, we suspect that for extremely deep networks (e.g. 1000 layers as in Figure 4C), curvature of the error landscape may be extremely problematic.  This means that the network is so sensitive to changes in the first layer that effective optimization of the 1000 layer network with a first-order optimization method is impossible.  Indeed, we set $g=1.05$ in Figure 4C precisely to deal with this issue.

Our experimental results show that even though depth did not clearly improve the training error, the initialization scheme was nevertheless effective at allowing training of these very deep networks to go forward.  Namely, almost all models with correctly chosen $g$ that were not broken, due to a mismatch of learning rate hyper-parameters to architecture, reached zero or near-zero training classification error or extremely low reconstruction error, regardless of depth.  Further research is necessary to determine whether or not more difficult or different tasks can make use of very deep feedforward networks in a way that is useful in applied settings.  Regardless, these results show that initializing very deep feedforward networks with Random Walk Initialization, $g$ set according to Figure 2 or as described in the section {\it Calculation of the Optimal $g$ Values}, as opposed to $g=1$, is an easily implemented, sensible default initialization.

\subsubsection*{Acknowledgments}

We thank Quoc Le and Ilya Sutskever for useful discussions.

\bibliography{infinit_iclr2015}
\bibliographystyle{iclr2015}

\end{document}